\documentclass[12pt, english, a4paper]{article}

\usepackage{epstopdf}
\usepackage{graphicx}
\graphicspath{ {./} }

\usepackage{subcaption}

\title{Parallel genetic algorithm for planning safe and optimal route for ship}
\author{Ivan Yanchin \and Oleg Petrov}
\date{Saint-Petersburg State Marine Technical University, Saint-Petersburg, Russia}

\begin{document}

\maketitle

\begin{abstract}
The paper represents an algorithm for planning safe and optimal routes for transport facilities with unrestricted movement direction that travel within areas with obstacles. Paper explains the algorithm using a ship as an example of such a transport facility. This paper also provides a survey of several existing solutions for the problem. The method employs an evolutionary algorithm to plan several locally optimal routes and a parallel genetic algorithm to create the final route by optimising the abovementioned set of routes. The routes are optimized against the arrival time, assuming that the optimal route is the route with the lowermost arrival time. It is also possible to apply additional restriction to the routes.

\textbf{keywords}: Genetic algorithms, maritime navigation, parallel computing, parallel genetic algorithms, autonomous transport.
\end{abstract}

\section{Introduction}
In the modern world we often hear about drones, self-driving cars and, in general, about increasing transport automation. The vast majority of research is dedicated to cars, however, cars are usually used for relatively short trips, generally taking several days at most, with breaks so that drivers can rest. It is almost impossible to move great amounts of goods from one continent to another by car or truck. Ships are used for that. For example, it usually takes about four or five weeks to travel from Brazil to Europe and it is possible to transport thousands of tonnes of cargo in a single voyage. In addition, it might be economically efficient to either reduce the crew size, through deeper automation or even develop unmanned ships that are able to function without on-board supervision~\cite{2017-cross-autonomous-ships-101}.

Such unmanned ships need to autonomously operate their internal systems and control their technical operations. Moreover, such a ship needs to navigate autonomously, which means that it should be able to plan its route without human assistance. In contrast with cars, ships travel in areas of almost infinite size, where it is usually possible to travel in almost any direction. As a result, we could not use methods, that are used to plan routes for autonomous cars, to solve the same task for autonomous ships, because the possible directions of a car are restricted.

Research shows that optimizing ship routes can also have economic benefits: an optimal route leads to lower fuel consumption and lowers operational costs~\cite{2018-grifoll-weather-ship-routing-economics}.

This paper describes an algorithm that could be used to plan routes for ships. The algorithm does not need any human attention while it operates, as a result it could be employed by a navigation system of an autonomous ship. The proposed algorithm tries to plot a safe and optimal route for a ship. Plotting an optimal route can be considered as an optimisation problem. A ship is a complex technical entity that is affected by several physical processes. Just like most of the vehicles, it is not able to turn on-place and it is not able to stop instantly. The turn diameter of a ship may be up to 4 times longer than the ship's length~\cite{2007-house-ship-handling}. A ship travels in water and as water is not dense, it creates relatively low resistance and inertia makes the ship move for a long time even after the engines are shut down. The distance a ship passes before it fully stops may exceed two nautical miles (about 3.7 kilometres). These distances are greater than a ship itself, and they affect the ship manoeuvring capabilities. Thus, these characteristics should be taken into consideration while planning a route for a ship.

The methods, used to optimize complex dynamic systems are commonly divided into two groups: strict methods and models, that are adapted to real-life interaction of the modelled objects with their environment; and methods of ``soft computing'' that employ intellectual technologies. Evolutionary models based on genetic algorithms have shown their effectiveness~\cite{2011-nechaev-catastrophe-theory} for multi-objective optimisation problems. Optimisation algorithms are discrete procedures that consist in modifying the initial set of solutions in order to find the best one. According to the dynamic catastrophe theory, application of such evolutionary methods to the development of the ``ship -- environment'' system is expressed by generating multiple alternate solutions and choosing the optimal one~\cite{2014-nechaev-ship-unsinkability}.

In case of the problem that consists in planning a route for a ship, the problem area may be an extremely large portion of water area. In addition, it is important to consider water currents while planning a route; moreover, the problem area changes with time because of tides that make depths vary periodically. Strict methods fail to consider these variations or require too much time to produce an acceptable solution~\cite{2005-krichevsky-intellectual-management}. Therefore, we think that heuristic methods may be considered a suitable solution for this problem.

Section 2 describes the task in more detail, and states some considerations about it. Section 3 provides an overview of existing solutions for the problem. Section 4 describes the proposed method in detail and section 5 provides experimental results. Section 6 describes topics that need further research and possible improvements for the method. Section 7 represents our findings and conclusions.

\section {Task}
Ocean ships transport millions of tonnes of cargo all over the seas. Ship fuel costs a lot and a ship consumes tonnes of fuel during a voyage. Fuel consumption may be reduced when the route is optimal, which means, the ship is moving with its optimal speed and does not manoeuvre intensively~\cite{2017-rodrigue-fuel-consumption-no-url}. In addition, cargo may have expiration dates so it is essential to deliver it in time. Ships usually travel in well-known areas where routes may be predefined. In some cases, a ship has to travel in an area that has not been studied profoundly, or, for some reasons, follow a unique route. For example, a scientific ship while doing its research has to travel within poorly studied areas. In some cases a ship needs to manoeuvre intensively even in well-studied areas, for example, at a highly loaded seaport when it travels to or from a pier. For instance, Europoort at Rotterdam serves thousands of ships per year~\cite{2017-rotterdam-busy-no-url}. It is difficult to manoeuvre safely when a port is overcrowded. If the action is taking place in difficult circumstances, for example in a strait with a complex seabed, the weather conditions become important. Sea currents and winds affect the ship's movements, therefore it is essential to consider the weather when planning a route for a ship.

Ships do not travel in void, so there is a water area where the action takes place. This area is restricted by coasts, even though it is very large. The bottom of such an area has a complex structure and there is a special kind of maps that describe this territory and that are used in maritime navigation. One of the important characteristics of a water area is a nautical chart that provides depth measurements at particular regions of an area. A ship is able to travel only within the areas that are deep enough. Nautical charts tend to provide average depths while the depths may change within a day because of tides. In some cases, a ship is able to pass through a particular area only during the rise of the sea level, because the area is too shallow or the ship sits in water too deeply. A ship has to wait until the level rises enough and avoid appearing in the area during the fall of the level; otherwise it will crash. The nautical chart provides information about the depths of particular regions and therefore about the obstacles that do not allow a ship to travel.

For this task, we assume that a route for a ship consists of several points called waypoints. The first waypoint of a route is called the start and the last point is called the destination. Part of a route between two adjacent waypoints is an edge. While following a route, a ship must sequentially arrive to every waypoint of the route. Every waypoint corresponds to a geographical location with particular coordinates. We assume that the area where the action is taking place is relatively small (tens of square nautical miles), as a result we may ignore the curvature of the Earth, use meters for coordinates instead of degrees, and use $x$ and $y$ axes instead of latitude and longitude. The proposed method can be adapted to larger areas and degree-based coordinates if needed. Every waypoint also has additional parameters that describe how a ship should follow the edge after the waypoint. Such parameters include the moment of time when a ship is expected to arrive to a waypoint, the moment of time, when a ship is expected to depart from a waypoint, the speed a ship is expected to move with while it follows the edge after the waypoint and, optionally, radius of the curve a ship must follow if a route requires it to turn.

Therefore, the task is to plan a safe and optimal route for a particular ship in a particular area starting at a predefined location and ending at a predefined location. A safe route is such a route that a ship is able to follow without crashing into an obstacle (like an island or area's seabed). An optimal route is such a route that is shorter in terms of travel time, than any other route that connects the start and destination points. We use time to compare routes instead of distances, because a ship may follow a longer route at a higher speed and arrive earlier than when it follows a shorter route at a lower speed.

This paper describes an algorithm that does not take the weather into account, ignores sea level changes and assumes that a ship travels alone, there are no other ships nearby.

\section{Related works}
Pathfinding is a well-known task that consists in generating a path from the start location to the destination. Different approaches to solve this task exist.

Paper~\cite{2018-grifoll-ship-weather-pathfinding} describes the task of planning ship route taking weather conditions into account. The paper uses A\textsuperscript{*} algorithm with heuristic and takes into account wind waves (speed and height). As a result, routes bypass areas with high waves that may cause danger. Paper~\cite{2018-grifoll-weather-ship-routing-economics} by the same authors assumes optimization of a route in terms of costs. The authors conclude, that optimizing a route is a multi-objective optimization problem, and thus a genetic algorithm can be used for it, but instead they use A\textsuperscript{*} algorithm with a heuristic that takes costs into account. Both papers use A\textsuperscript{*} algorithm which requires a graph or a grid that describes area where the action is taking place and wave information must also be a part of this grid. In both papers wave information is provided by external systems and is assumed to be always available.

Paper~\cite{2012-shao-dynamic-programming-ship-route} describes a dynamic programming method for plotting a route for a ship. In this case route is a sequence of tuples each of which describes engine power and ship's heading during voyage. This method requires hydrodynamic model of a ship for which a route is being planned and method's precision depends on the precision of the model. Moreover, this method assumes that an external optimization method is used to solve the dynamic programming problem.

Paper~\cite{2015-simonsen-state-of-the-art-weather-routing} describes state-of-the-art in field of route planning, providing survey of existing methods. In case of dynamic programming methods, a graph or a grid is required, and method's effectiveness depends on resolution of a grid, in some cases it is even possible to miss an obstacle that affects the route's safety. In case of conventional graph-based approaches, like Dijkstra's algorithm~\cite{1959-dijkstra-algorithm}, algorithm requires a graph to be preliminary constructed. In general, only single-objective route optimization is possible using graphs, but multi-objective route optimisation can be achieved through synthetic edge cost that is computed taking all objectives into account. Genetic algorithms are also used for route planning, moreover, genetic algorithms can be used for multi-objective route optimization. However, the solution depends on the quality of the initial generation. Authors also provide their own solution to the problem called DIRECT. DIRECT method is a multi-objective route optimization method and represents the solution space as an $n$-dimensional hypercube, where $n$ is the number of objectives. The method consists in dividing this hypercube to several hyperrectangles according to objectives. This method does not require a graph or a grid. However, even though the method can be used for multi-objective optimization, in case of multiple objectives, the dimensionality of the hypercube also increases, and the performance degrades. Moreover, the performance degrades when the number of waypoints increases.

Paper~\cite{2004-hinnenthal-ship-routing-study-ga} describes an algorithm that can be used to optimize a ship's route. This method optimizes the route in terms of arrival time and fuel consumption, and requires graph. The described method represents a route as a B-spline, and route optimization consists in optimization of location of spline vertices. Authors compare a simplex method and a genetic algorithm. Genetic algorithm showed better result and managed to find Pareto-optimal solution, while the simplex method finds the local optimum faster.

Therefore, using a genetic algorithm to solve the described task is a considerable option. Genetic algorithms can find optimal solutions when strict mathematical methods are not applicable~\cite{2005-krichevsky-intellectual-management}. It should be noted that genetic algorithms do not guarantee that the produced solution is globally optimal; it is only the most optimal solution between the others handled by an algorithm. However, in some cases genetic algorithms are able to find an acceptable solution much faster~\cite{2005-krichevsky-intellectual-management}.

Paper~\cite{2008-kosmas-ship-route-hybrid-ga} describes almost the same task as this paper does. The authors use hybrid Genetic Algorithm to plot a route for a ship from point A to point B. They recommend to initially plot a straight route between the points and then tweak it in such a way that it avoids obstacles. There are several waypoints within a route. The authors of that paper use constant amount of 20 waypoints for all routes. The authors recommend rotating coordinate axis in such a way that the line that connects the first and the last waypoints is parallel to the $x$ axis, as a result, the obstacle avoidance is achieved through modifying the $y$ coordinate of a waypoint keeping the $x$ coordinate constant. A genetic algorithm is used to modify $y$ coordinates of waypoints. The proposed algorithm allows routes that cross an obstacle, so, in general, it might be possible that the final route produced by the algorithm could not be followed because it crosses an obstacle. We think that this algorithm is not suitable for the task described in this paper because to our consideration an algorithm must either produce a safe route or explicitly indicate that it is not possible, we need an algorithm to take the ship dimensions into account. Moreover, this algorithm requires a preliminary obstacle detection in order to make it possible to check whether a route crosses them. It describes obstacles as self-contained entities, not through depths, as a result it might not be possible to dynamically model obstacle size changes as a result of depth changes caused by tides. Later these authors published another paper,~\cite{2012-kosmas-simulated-annealing-ship-route} dedicated to the same problem. In the newer paper they decided to use the simulated annealing method instead of a genetic algorithm. The newer method improves findings of the earlier paper, but has the same drawbacks: it requires preliminary obstacle detection and describes obstacles as objects, not using nautical charts.

Sometimes it is relatively difficult to find a solution for a problem using a simple genetic algorithm because it may require too much time to find a solution, when the problem area it extremely large or the individual itself is a relatively complex object. In this case, one may try to employ a parallel genetic algorithm that reduces the required time or allows processing more individuals within the same time period. Nowadays it is easy to get access to powerful computational facilities that allow for massive parallel data processing, and thus parallel genetic algorithms have become widely used~\cite{2016-salza-parallel-ga-cloud-containers, 2013-ferrucci-hadoop-ga, 2017-ferrucci-hadoop-map-reduce-parallel-ga}.

At~\cite{2009-Hassani-standard-parallel-ga} the authors describe standard approaches to genetic algorithms and possible improvements to them through parallelisation. In case of a parallel genetic algorithm, multiple populations evolve in parallel spreading computations to multiple processors, and thus increasing performance. Genetic algorithms allow to produce an acceptable solution in just a few iterations and further processing improves it except when the algorithm converges to a local optimum. The authors of that paper also describe basic genetic operators and their variations. They also describe several common parallelisation techniques for genetic algorithms. Additional attention is paid to Dual Species Genetic Algorithm, where the entire population is divided into two subpopulations. The authors of that paper conclude that genetic algorithms are able to produce an optimal solution or a good estimate if the algorithm is properly tweaked for the task, which is a relatively hard task. The most important difficulty related to genetic algorithms is a premature convergence to a local optimum. We discuss methods that may help to avoid this.

\begin{figure}[hbt]
    \centering
    \includegraphics[scale=1]{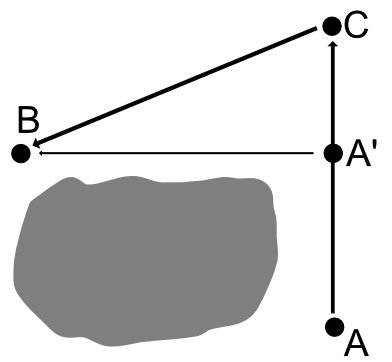}
    \caption{The ship is not able to travel directly from the point~$A$ to the point~$B$, thus it has to travel firstly to point~$C$ and then to point~$B$. However, route $A$-$A'$-$B$ is better, but it is not allowed due to the grid resolution.}
    \label{img:grid-restriction}
\end{figure}

It is a quite common approach to represent the depth map as a grid, and the ship is expected to travel by its edges and change direction only at nodes. However, if the grid resolution is relatively high, which means that nodes are placed close to each other, then complexity of the pathfinding algorithm that is used to find a route increases. If the grid resolution is relatively low, which means that nodes are placed sparsely, then the complexity of the algorithm becomes lower, but in this case the ship's manoeuvring capabilities are heavily restricted by the grid and thus it may miss a better option (figure~\ref{img:grid-restriction}). Therefore, it may be worth it to avoid using a grid or a graph and use a continuous map instead of discrete one.

\section{Method}
The task is to plan a route for a ship so that the route is as short as possible in terms of travel time, and does not make a ship to collide with an obstacle. Even though weather and, especially, speed and direction of currents affect ship movements, we will not discuss this in this paper. In addition, for the sake of simplicity and in order to concentrate on route planning, we assume that there are no tides and depths stay constant all the time. As inputs, the algorithm accepts the description of a ship that needs a route, a map of an area where the action is taking place, and the coordinates of the start and the destination points. The map is represented as a depth map, providing information about average depths of the area. Within the entire algorithm and related models, coordinates are expressed in metres, not in degrees and the area is assumed to be plain, without curvature.

\subsection{Ship}
A ship is a complex craft that travels waterways. It is important that a ship travels in areas deep enough otherwise it is in danger. A shipwreck affects surrounding environment (due to fuel spills), destroys property (cargo and the ship itself), kills people, and, in some cases, blocks waterways, preventing other ships from travelling. Obviously, shipwrecks should be avoided at all costs. We will not discuss different shipwreck reasons in this paper; we will only concentrate on depths. For safe passage, a ship needs waterways along its route to be sufficiently deep. It is required that the depth of an area is greater than the draught of the ship. Even though a ship has other dimensions (length, beam and so on), this paper mostly ignores them, and we assume that the only danger comes from depths. Therefore, the system that plans routes must consider draught and must not make a ship travel at shallow areas. Water currents may also affect depths, tides make depths vary during a day, and if a particular area is deep enough during a tide, it may be too shallow during an ebb. An ideal route planning system should also consider tides while doing its job. For this paper we assume that depths are constant and do not change over time, the algorithm that takes tides and currents into account is a subject of the further research.

\subsection{Two-algorithm solution}
The proposed method is based on two subalgorithms each of which solves a particular task. The first subalgorithm plans a single route while the second one takes several routes produced by the first one as input and produces a single one based on the inputs. Both subalgorithms are heuristic. The first subalgorithm is a producer that is used by the second one when it needs routes for its operation. The second subalgorithm is a parallel genetic algorithm that treats routes as individuals and generates the solution from the initial population generated by the first subalgorithm.

\subsection{Planning a route}\label{sct:planning-route}
The first subalgorithm plots a single route from the start to the destination point. We propose to use a heuristic method that has a lot in common with traditional genetic algorithms but without mutations and crossover.

The algorithm is iterative and assembles routes from separate points. The process starts at the start point. During a single iteration, one new point is selected and appended to the route that is being constructed. The process stops when the route reaches the destination or when the total amount of points becomes greater than a boundary. The boundary restricts the total amount of points in a route and prevents infinite loops that may occur if the algorithm fails to find a route to the destination. The value of the boundary depends on the length of the ship and the size of the area where the action is taking place.

\begin{figure}[hbt]
    \centering
    \includegraphics[scale=0.2]{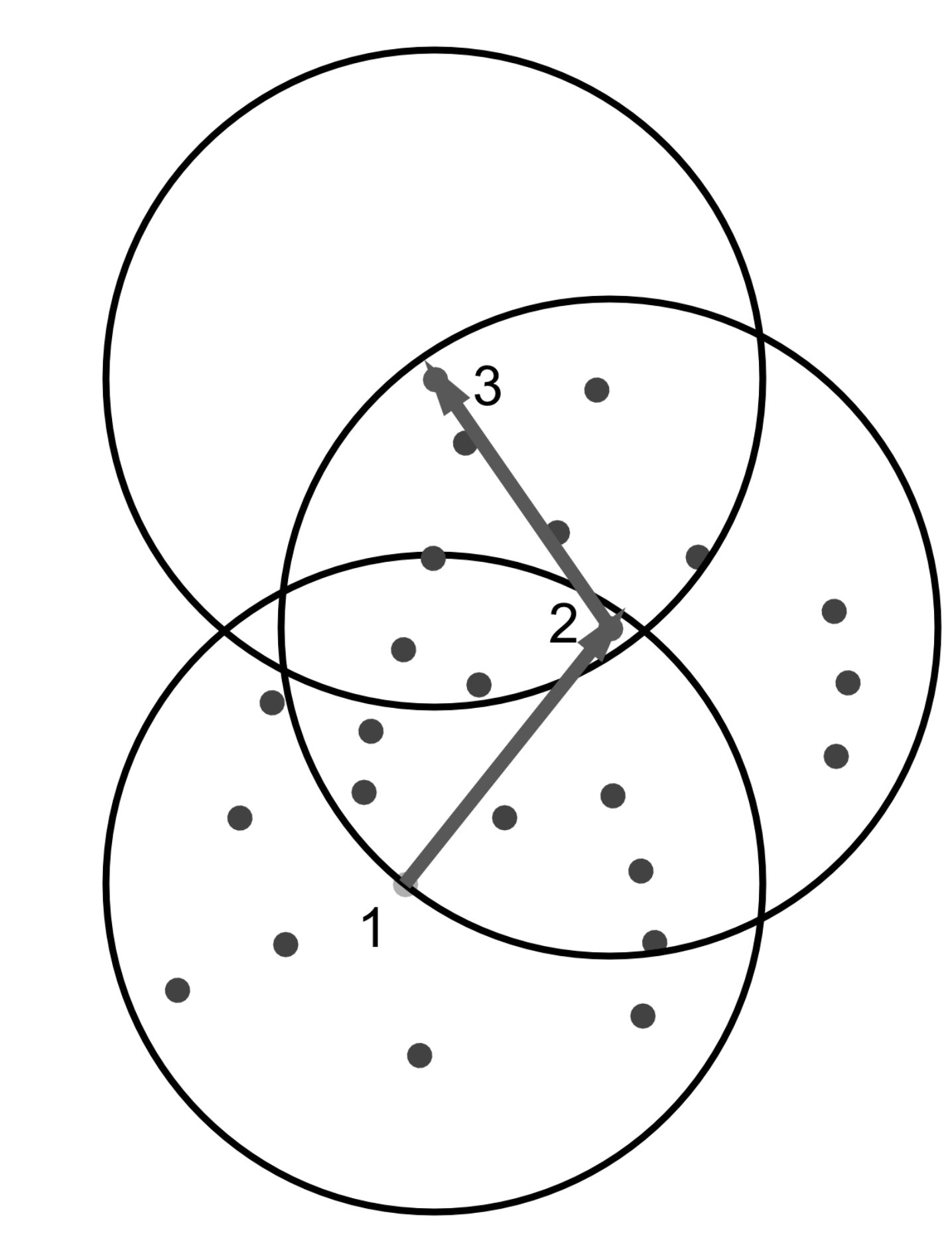}
    \caption{Route planning illustration.}
    \label{img:route_planning}
\end{figure}

Figure~\ref{img:route_planning} illustrates the method. On the first step, several points are created and then the fitting function (described later) assesses all of them. The best point (with the greatest fitting value) is appended to the route, and becomes its new last point. Then a new step begins. Once the route reaches the destination, the process stops. The points that are generated at every step are random and it is difficult to predict when the point that is equal to the destination point is generated.

At every step, random points are generated within a circular area, whose radius depends on the size of the ship for which the route is constructed (say, 5 times longer than the ship's length). We call this area ``domain of a ship''. A point can be generated anywhere within the domain, but could not be outside. If the destination point is located within the domain of the last point of a route and it is possible to travel between these points, the destination point is appended to the route without any assessment.

Generally, the size of the domain stays the same during the entire planning process, but we also would like to present three additional methods to vary it.

\begin{figure}[hbt]
    \centering
    \includegraphics[scale=0.3]{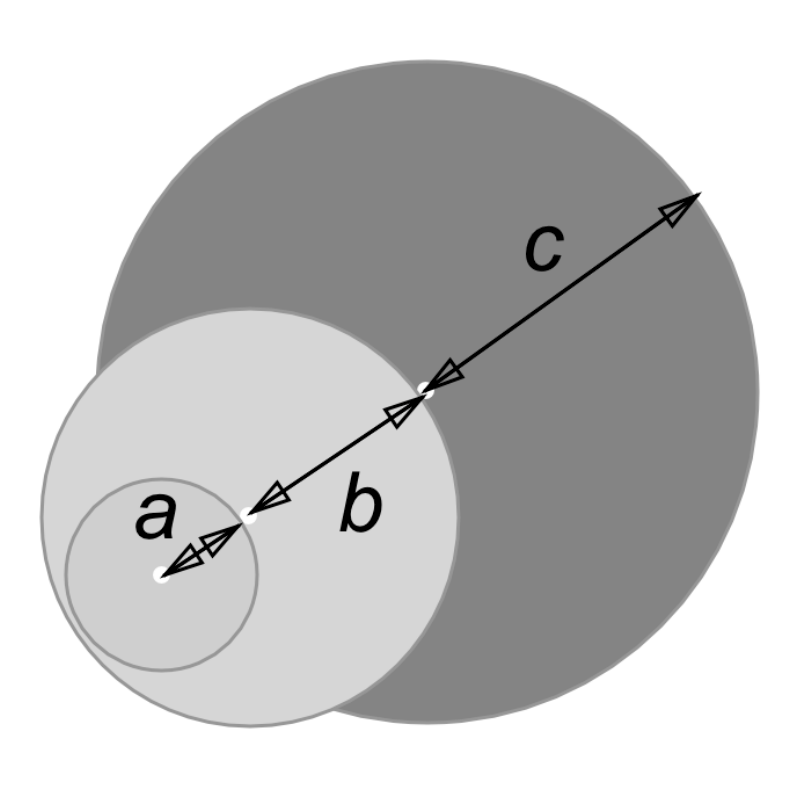}
    \caption{Domain size grows sequentially.}
    \label{img:grow_dist}
\end{figure}

The first variant assumes that the radius of the domain grows from step to step. Figure~\ref{img:grow_dist} illustrates this method. At the first step the radius is $a$, at the second step it is $b$, and at the third step it is $c$, given $a < b < c$. The radius changes between the steps and never changes during a step, so that for all points that have been created during a step the radius stays the same.

\begin{figure}[hbt]
    \centering
    \includegraphics[scale=0.3]{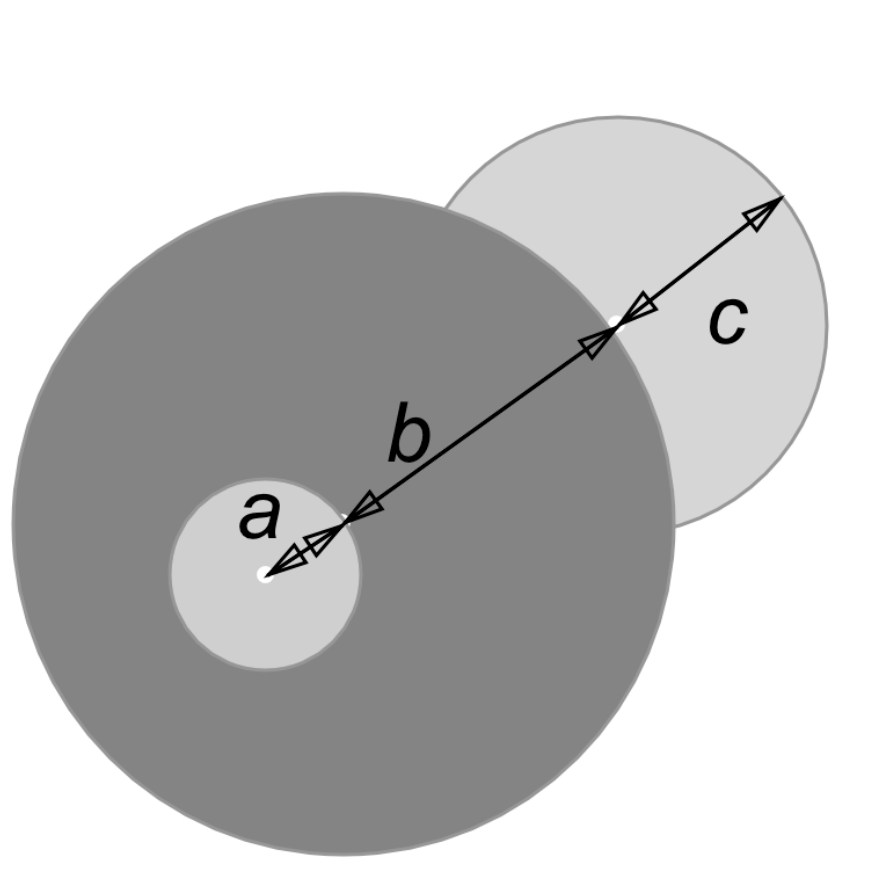}
    \caption{Domain size changes randomly.}
    \label{img:random_dist}
\end{figure}

The second variant assumes that the radius of the domain changes randomly from step to step. On figure~\ref{img:random_dist}, at the first step the radius is $a$ and at the second step the radius has much greater length, $b$, at the third step the radius $c$ is smaller than at the previous one. Like in the first variant, the radius is the same for all points of a step.

\begin{figure}[hbt]
    \centering
    \includegraphics[scale=0.4]{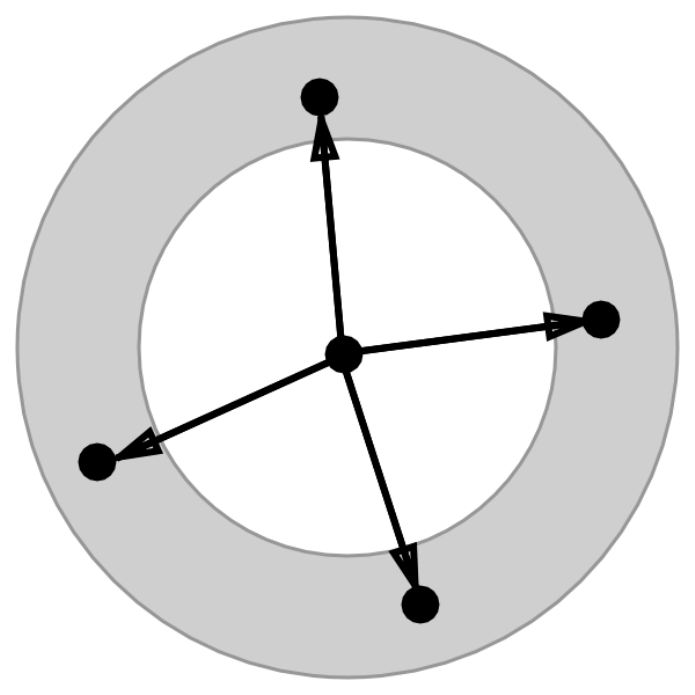}
    \caption{Only painted area may contain points.}
    \label{img:min_dist}
\end{figure}

The third variant does not assume any systematic changes, but it introduces a minimal domain radius. On figure~\ref{img:min_dist} points are generated only within painted area, and there must be no points within the blank area. This variant is intended to make such routes that contain fewer waypoints and thus encourage large and straight route edges.

\subsubsection{Sector-based point creation}
\begin{figure}[hbt]
    \centering
    \includegraphics[scale=0.4]{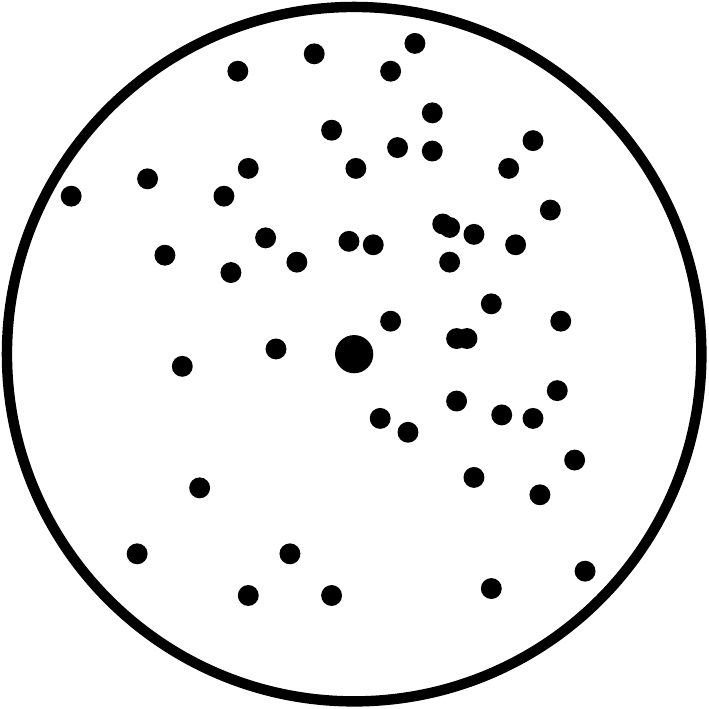}
    \caption{Bottom-left area contains less points than others.}
    \label{img:uneven_points}
\end{figure}

The point generation method described above works fine when the total amount of points is relatively great. When there are many points, they evenly cover the domain and almost every area is examined. When the amount of points is relatively small, there might be areas that contain no points, and thus that will not be examined. Figure~\ref{img:uneven_points} illustrates this issue: the bottom-left side of the domain contains fewer points than the others, so it would not be examined properly, and the algorithm may miss an optimal point.

\begin{figure}[hbt]
    \centering
    \includegraphics[scale=0.4]{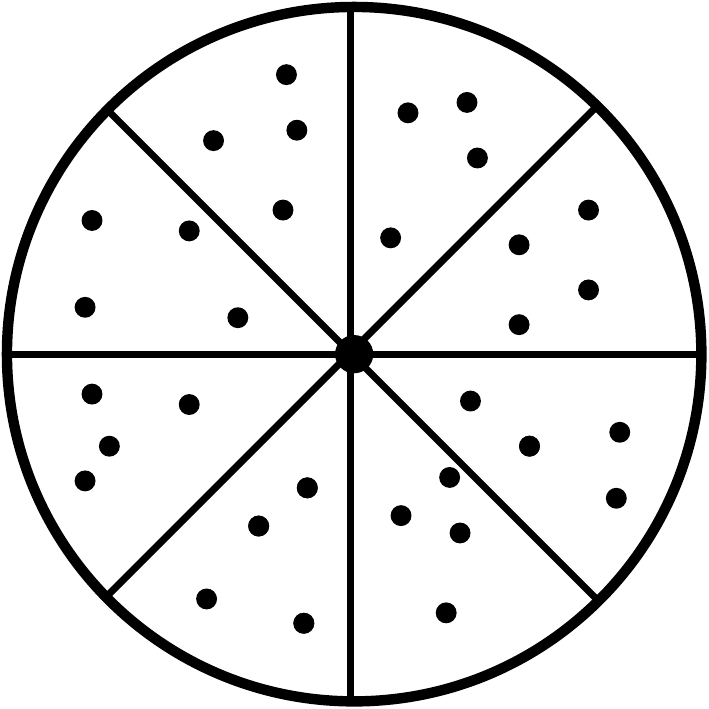}
    \caption{All sectors contain the same amount of points.}
    \label{img:sectors}
\end{figure}

In order to overcome this we recommend splitting the entire domain to several sectors of equal sizes, and deal with each sector independently. All the sectors contain the same amount of points. The points within a sector are distributed randomly, but all the sectors contain the same amount of points (figure~\ref{img:sectors}). The location of a point within a sector
is random, only the sector itself is predefined.

As a result, the points are evenly distributed over the domain and all its areas are examined. Experiments have demonstrated that using sectors we can reduce the total amount of points without any reduction of route quality. This means that the sector-based approach ensures that points are distributed in such a way, that we need fewer points in order to find the optimal one. In theory, it is possible that the points within a sector localize around the centre of the domain, which makes the areas close to the edges contain fewer points. However, experiments have shown that this is not an issue when a proper random number generator is used, especially, uniform distribution.

\subsubsection{Fitting function for a point}
Effectiveness of the algorithm that plans separate routes depends on the function used to select the best point on every step. This function is expected to select such a point that keeps the route optimal and safe but also reduces the distance left until the destination.

\begin{figure}[hbt]
    \centering
    \includegraphics[scale=0.3]{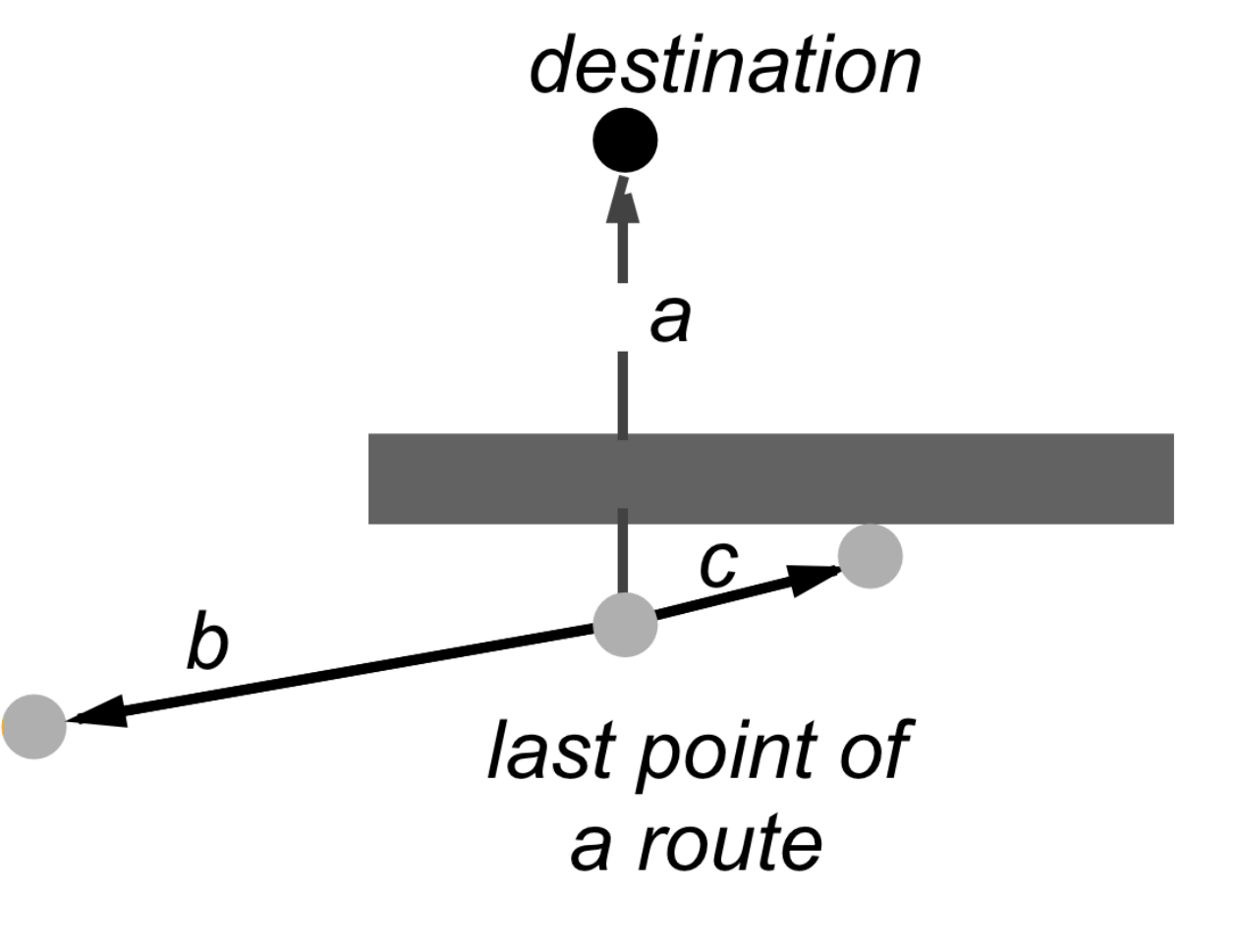}
    \caption{Way $b$ will never be selected.}
    \label{img:move_fwd_imp}
\end{figure}

During the research, we have examined several fitting functions for points. The initial variant was to compare distances from a point to the destination and choosing the point with the shortest distance. In order to make shorter distances receive greater fitting, the distances were negated. This method seemed to be good enough, it used to make points to have greater fitting when they were closer to the destination. During experiments, we have found out that this method does not allow a ship to move backwards, which may be required if moving forward is not possible. For example, in Figure~\ref{img:move_fwd_imp}, Way $a$ is blocked, but Way $b$ will always receive low fitting because the distance to the destination point is larger than for Way $c$ which will eventually be blocked. In order to overcome this we decided to allow a ship to move backward when it is not able to move forward.

We created a two-component fitting function~\ref{eq:two_comp_ff}. The value of this function is a sum of two components, the first assesses how close the point is to the destination and the second one indicates in what degree the ship moves back. Coefficients $\xi$ and $\psi$ are used as weights for the components, usually they both are equal to $1$, but could be adjusted in order to tweak the function, making them smaller or greater than $1$ means making the corresponding component of the equation proportionally less or more important.

\begin{equation} \label{eq:two_comp_ff}
    FF = -\xi\frac{d_{td}}{d_{cd}} + \psi\frac{\Theta}{180}
\end{equation}

Value $d_{td}$ represents the distance between the point that is being examined and the destination. Value $d_{cd}$ is the distance between the last point of the route and the destination. As a result, point fitting depends on the degree of how close the point that is under examation is to the destination compared to the last point of the route. In order to indicate that shorter distance is better, we use a minus sign in front of it, and make values with smaller magnitude become greater.

\begin{figure}[hbt]
    \centering
    \includegraphics[scale=0.3]{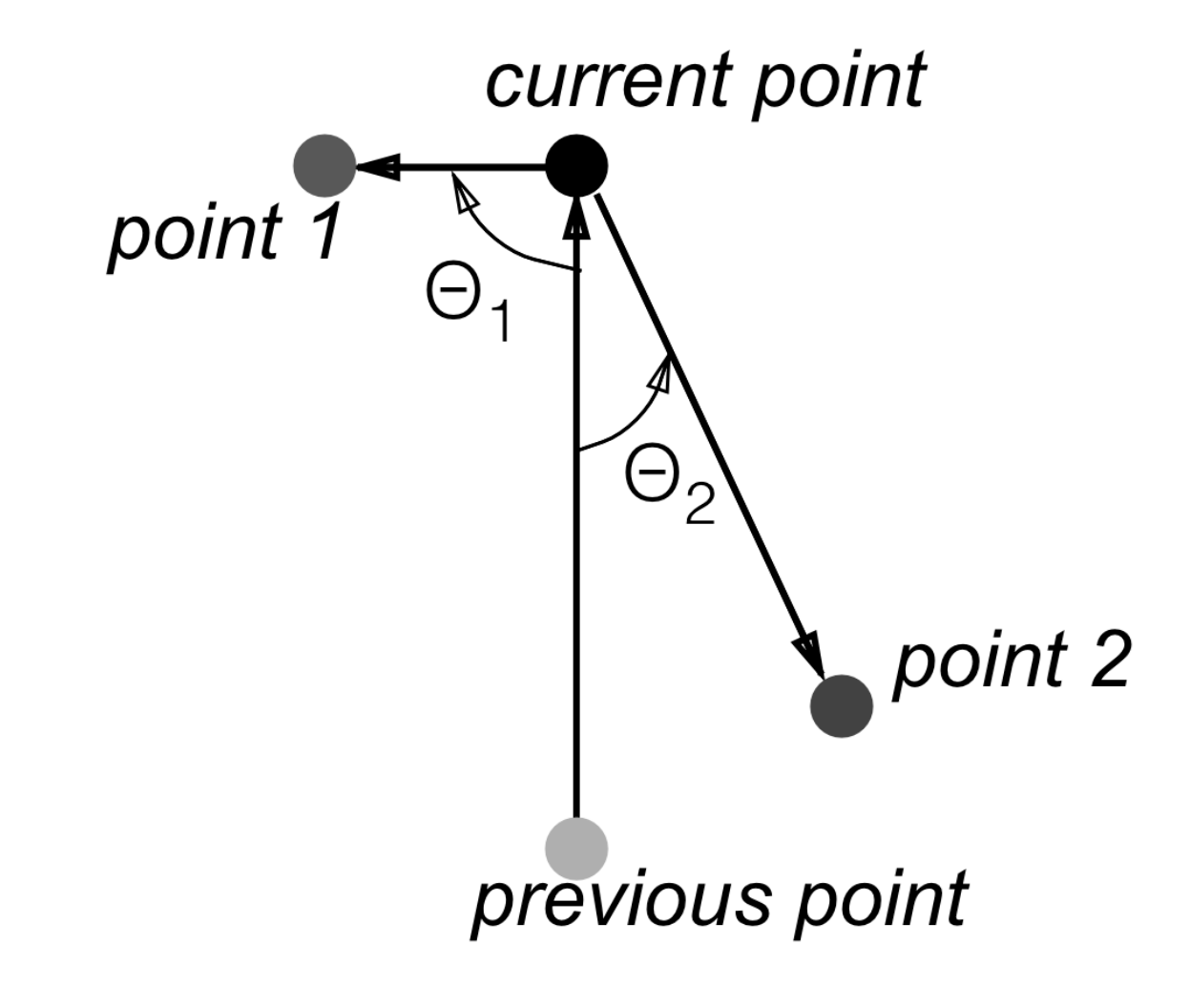}
    \caption{The wider angle means the ship is moving forward.}
    \label{img:move_back_edge}
\end{figure}

The move backward degree is computed using the angle between the edge that is between the point that is being examined and the destination, and the edge between the last point of the route and the point before the last one (figure~\ref{img:move_back_edge}). This angle is represented with $\Theta$ ($\Theta_1$ and $\Theta_2$ on figure~\ref{img:move_back_edge} denote the same property of different points). We represent this angle in degrees and divide by 180 in order to normalise it, so that the value is within the $[0, 1]$ range.

We use an angle instead of the distance between the point that is being examined and the point before the current one because it is not easy to decide what to compare this distance with in order to understand whether a ship will move backwards or not. We may try to compare this distance to the one between the current point and the point before it. On the other hand, we could also easily imagine a pair of points where the point that makes the ship move forward is close to the last point of a route, and the point that makes it to move backward is far from the last one. In this case, the more distant point should be selected, and this is the point that makes the ship turn back.

This two-component fitting function makes the ship move forward as long as possible, but when the way is blocked, it allows turning back. The penalty for turning back prevents the ship from turning back when not actually needed and doing so repeatedly, moving between a pair of points. All points that make a ship change its direction to the opposite have lower fitting than any point that makes it keep its current direction. When it is not possible to move forward, all points that make a ship move like that receive the lowermost possible fitting, and moving backward becomes an acceptable solution. We use the term ``to move forward'' to denote situations when a ship keeps its current direction and the term ``to move backward'' for situations when it changes its direction to the opposite one. As a result, the forward direction may actually make a ship increase distance to the destination (if a ship is heading in the opposite direction), but we encourage a ship to preserve its direction as long as possible, regardless where it actually moves.

\subsection{Genetic algorithm for the final route}
We use a genetic algorithm to plan the result route. The routes produced by the first subalgorithm form the initial population. When the genetic algorithm needs additional individuals, it uses the first subalgorithm to create them. For example, when this genetic algorithm selects individuals that are to form the next generation, it may find out that the number of these individuals is less than the adjustable boundary, in this case the algorithm uses the first subalgorithm to create several new routes; this may happen at any iteration of the genetic algorithm, not only while creating the initial generation. The number of routes to generate and the value of the boundary that makes the algorithm generate additional routes are configurable and may vary from one run to another.

The genetic algorithm uses routes as individuals and different routes may have different amount of waypoints which means that individuals have variable size. Variable-length genetic algorithms have shown their effectiveness for the planning~\cite{2005-brie-genetic-planning-var-len} and pathfinding tasks\cite{2016-cekmez-multi-auv-path-planning-parallel-ga-cuda} and for the tasks where restricted genome size leads to restricted choice of possible solutions which, in turns, reduces the overall effectiveness of the algorithm~\cite{2000-lee-variable-len-evol-alg}.

Like any other genetic algorithm, the one being discussed uses mutation and crossover operators to create new individuals and to modify existing individuals in-place, respectively. Every individual has a chance to mutate and take part in crossover with another one. Mutation and crossover are independent, so an individual can take part in both operations during a single step.

Three possible mutations could be applied to an individual. In case of the first variant, the mutation operator inserts a new, random waypoint at a random place of a route. This point is located within the domain of a randomly selected point that already belongs to a route. This operation allows avoiding relatively small obstacles making a route to be distant from them. In case of the second variant, we change the location of an existing point. The new location of the point is chosen within the domain around its original location. This variant also makes the route avoid obstacles but in contrast to the first variant, the total amount of points in the route stays the same. In case of the third mutation variant, the mutation operator removes a random point from the route and directly connects points before and after. This mutation makes the route simpler, removing points that do not affect its safety. When a point is selected for mutation, a random mutation variant is applied to it. All mutation variants have equal chances to be applied. The start point and the destination could not be mutated in any way. In some cases it is possible to get an invalid route after mutation, this is not generally an issue because such a route has lower fitting and thus, would not be selected for the next generation.

When several individuals are selected for crossover, they are grouped together and the crossover operator is applied to each group. The size of a group depends on how many individuals the crossover operator requires. Here we use only two individuals. Paper~\cite{2016-qiongbing-crossover-var-len-ga-path} provides research on possible crossover operators for tasks where solutions could be represented as chains of objects: in this case traditional one-point crossover operator is not effective, because it requires two solutions to have common parts, which might not be a very frequent case. This paper presents two crossover operators, that are generalization of the one proposed in the paper~\cite{2016-qiongbing-crossover-var-len-ga-path} for the cases when there is not graph. We call the first variant ``short-distance'' and the second one ``long-distance'' crossover. In case of the short-distance crossover we choose a pair of points that belong to different routes and for which the distance in-between is the shortest. In other words, we search for the points where the two routes are as close to each other as possible. In case of the long-distance crossover, we also choose a point within every route, but we do not restrict the distance between them, we simply pick a random point from a route. Then both routes are divided into two parts, the first one ends at the selected point and the second one starts after it. The routes exchange their second parts. As a result, the crossover operator creates two new routes, and the source routes stay the same. The difference between the two methods consists in the way we select the splitting point. Any two routes always have two equal points: the start one and the destination, and these are the points that could be selected by the first crossover method, because the distance between the routes at these points is zero. Crossover at these points is useless so, in order to avoid this, we decided never to choose the start and the destination points as splitting points.

The aim of the genetic algorithm is to select (or create) the most optimal and safest route among the others available. In order to select such a route, it uses the fitting function that examines every route and the better the route, the higher score the function assigns to it. Therefore, the aim of the algorithm is to maximise the score of routes. The genetic algorithm that is described here is expected to encourage safe and optimal routes, and thus, it needs such a fitting function, that assigns greater score to more optimal routes and discourages unsafe routes. The fitting function that is used for that uses the time when the ship arrives to the destination point. We negate the time in order to indicate that lower time is better. We use the arrival time instead of the route length because a ship may move faster and follow a longer route in less time than it needs to follow a shorter route with lower speed. A route that could not be followed because of an obstacle receives the lowermost possible score.

\begin{figure}[hbt]
    \centering
    \includegraphics[scale=1]{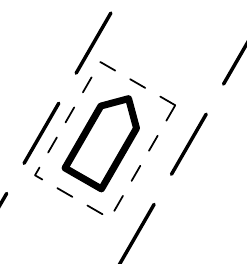}
    \caption{Thin dashed lines indicate the rectangle around the ship that is used to check whether the depth below the ship is sufficient. Using larger rectangle allows to ensure a channel (bold dashed lines) is wide enough.}
    \label{img:safe-area}
\end{figure}

This fitting function also ensures that a ship is able to follow the edges of a route. When the fitting function checks depths of the points by the edge it checks the depth not of a single point, but of a rectangle with dimensions slightly greater than the dimensions of a ship. This rectangle is rotated in such a way that its turn angle is the same as the turn angle of a ship. Figure~\ref{img:safe-area} illustrates this. Therefore, we ensure that a route does not make a ship pass through a channel that is too narrow by validating depths not of a waypoint itself, but of an area around to it. We use a slightly bigger rectangle in order to have some extra space on either side of a ship. We also require the area to have depths a bit greater than the draught of a ship for safety reasons.

The fitting function could estimate the moment of time when a ship is going to reach a waypoint so that it can use a kind of depth prediction facility to get expected depths of an area around a waypoint at this particular moment. We would not discuss depth prediction in this paper, but the proposed algorithm is able to handle areas that are safe for a ship only during a tide and effectively avoid these areas during an ebb in case depth predictions are available.

As one can see, in order to examine routes properly we need to know when a ship will arrive to the destination if it follows a particular route. In order to estimate the arrival time at the destination point, we need to know the arrival and departure time for all the points of the route. As we have mentioned earlier, every point of a route has additional properties. The properties of a point provide the moment of time when a ship is expected to arrive at it, the moment of time a ship is expected to depart, the speed it is expected to move with and the radius of the curve a ship has to follow if a route requires it to turn. When a new route is created or an existing one is modified we need to compute these properties again, because when the structure of a route changes the arrival and the departure time are invalidated. Moreover, as the properties of the point depend on the properties of a point before it, we have to update all the points, not only the modified ones. Using these properties, we are able to assign the score to routes and then choose the most optimal one.

\subsection{Planning a route in parallel}
Routes are relatively large entities, so it may be impossible to store many of them in memory at the same time. Another issue of genetic algorithms is preliminary convergence to a subset of solutions~\cite{2009-Hassani-standard-parallel-ga}. This means there is no significant difference between solutions, which leads to lower effectiveness of the genetic algorithm or even to inability to locate the global optimum. In order to overcome these issues, we propose to use a parallel genetic algorithm. Parallel genetic algorithms increase diversity of individuals because separate algorithms that are running in parallel may be configured differently. Therefore, it is possible to reduce the size of a single generation because the required diversity is achieved through different configurations of algorithms and not through the total amount of individuals\cite{2016-yanchin-parallel-ga-path-ru-en}.

Parallel genetic algorithms allow to find an acceptable solution faster and improve it within subsequent algorithm iterations~\cite{2009-Hassani-standard-parallel-ga}. Multicore central processing units (CPUs) and multiprocessor computers have become widespread recently. Utilising these facilities to solve the problem sounds reasonable because this allows computing parts of the solution in parallel, which in turn can also reduce the overall time consumption. We recommend considering a parallel genetic algorithm to solve the task.

In contrast to~\cite{2009-Hassani-standard-parallel-ga} where the authors propose to divide the entire population into two subpopulations and the only difference between the subpopulations consists in the methods used to choose individuals for crossover and mutation, we propose to make almost every aspect of an algorithm configurable. Different algorithms running in parallel may have different generation sizes, different crossover and mutation rates, different selection mechanisms, and different methods used to generate new individuals. The only two aspects that should be the same are the number of generations, in order to simplify synchronisation, and the fitting function, in order to make the individuals that belong to different algorithms be comparable.

Therefore, a set of independent genetic algorithms that are running at the same time in parallel may have different properties and as a result different sets of individuals within the generations. In order to improve the quality of the entire system, these algorithms could also be crossed over in order to modify their current generations and introduce new individuals with properties that could be inaccessible for a particular algorithm because of its configuration. In this case, all algorithms are randomly grouped into pairs and elements of a pair cross their current generations. When two algorithms are to be crossed over, the crossover operator that belongs to the first one is used and the individuals created during the operation are added to its current generation; the second algorithm acts as a ``donor'', only providing its individuals. The first algorithm of such a pair is chosen sequentially while the second one is random; as a result, all algorithms receive new individuals being the first of a pair. Such ``cross-algorithm crossover'' spreads diversity over the system. In~\cite{2009-Hassani-standard-parallel-ga} ways the two parallel algorithms may interact with each other are also discussed, but authors only allow several individuals to migrate from one subpopulation to another without any changes, while we recommend employing crossover because this allows producing new individuals for every algorithm which increases diversity\cite{2016-yanchin-parallel-ga-path-ru-en}.

Earlier we have discussed several ways to create new waypoints. In case of non-parallel genetic algorithm one have to choose a single way new individuals are created even though there are multiple methods available. A parallel genetic algorithm allows to use all possible variants. All these methods could be used at the same time by different algorithms that are running together in parallel, which also increases solution diversity and prevents generation convergence.

This approach is also known as island model, which means that several independent genetic algorithms with different properties are evolving independently like several populations that live on different islands do\cite{2015-skobtsov-evolutionary-computations}.

\section{Experimental results}
We have conducted several experiments to test the proposed method. We used synthetic depth maps for testing purposes. Synthetic data, firstly, allows us to prepare such maps that make the algorithm demonstrate all its capabilities, secondly, gives us full control over the map content and, thirdly, synthetic maps allow modelling difficult situations and seeing how the algorithm can handle them. We used such maps for benchmarking, verification tests conducted to ensure that the implementation is correct, and to better tune this implementation using verification results.

For the testing purposes we have implemented an application in C++ that performs route planning using the proposed algorithm. This application is a command-line application that runs under UNIX operating systems. The proposed method employs island genetic algorithm, and thus this application uses operating system threads to reach parallel execution of several genetic algorithms. It also performs crossover between islands in parallel using the same threads. We didn’t use any additional low-level optimisation techniques.

We have chosen three synthetic maps for benchmarking. Two of them are relatively small and are intended to demonstrate how the proposed algorithm handles difficult situations and edge cases. The third map is relatively large and contains many small obstacles that represent islands or reefs. On all maps point \textit{A} is the start point, and point \textit{B} denotes the destination location.

\begin{figure*}[h!t]
    \label{fig:bench-maps}
    \begin{subfigure}[b]{0.3\textwidth}
        \includegraphics[width=4cm]{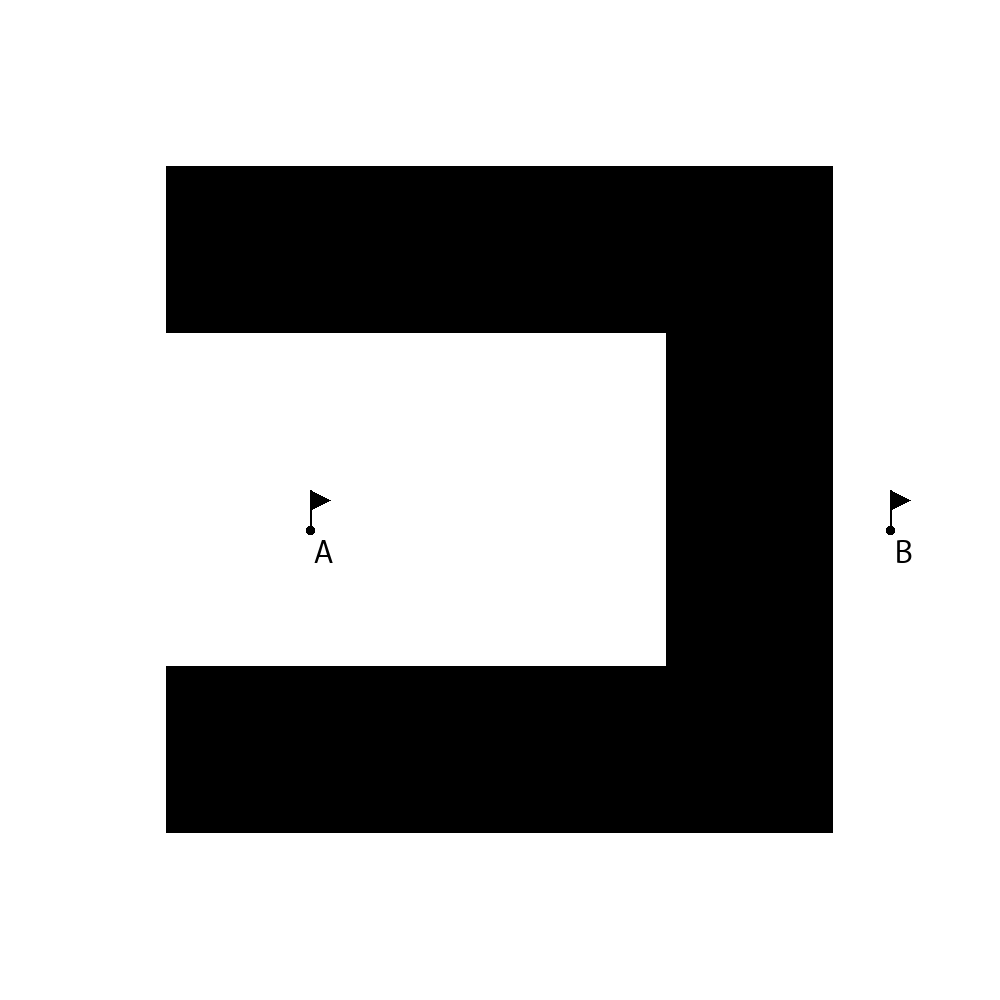}
        \caption{The first simple map.}
        \label{fig:move-back-test}
    \end{subfigure}
    \hfill 
    \begin{subfigure}[b]{0.3\textwidth}
        \includegraphics[width=4cm]{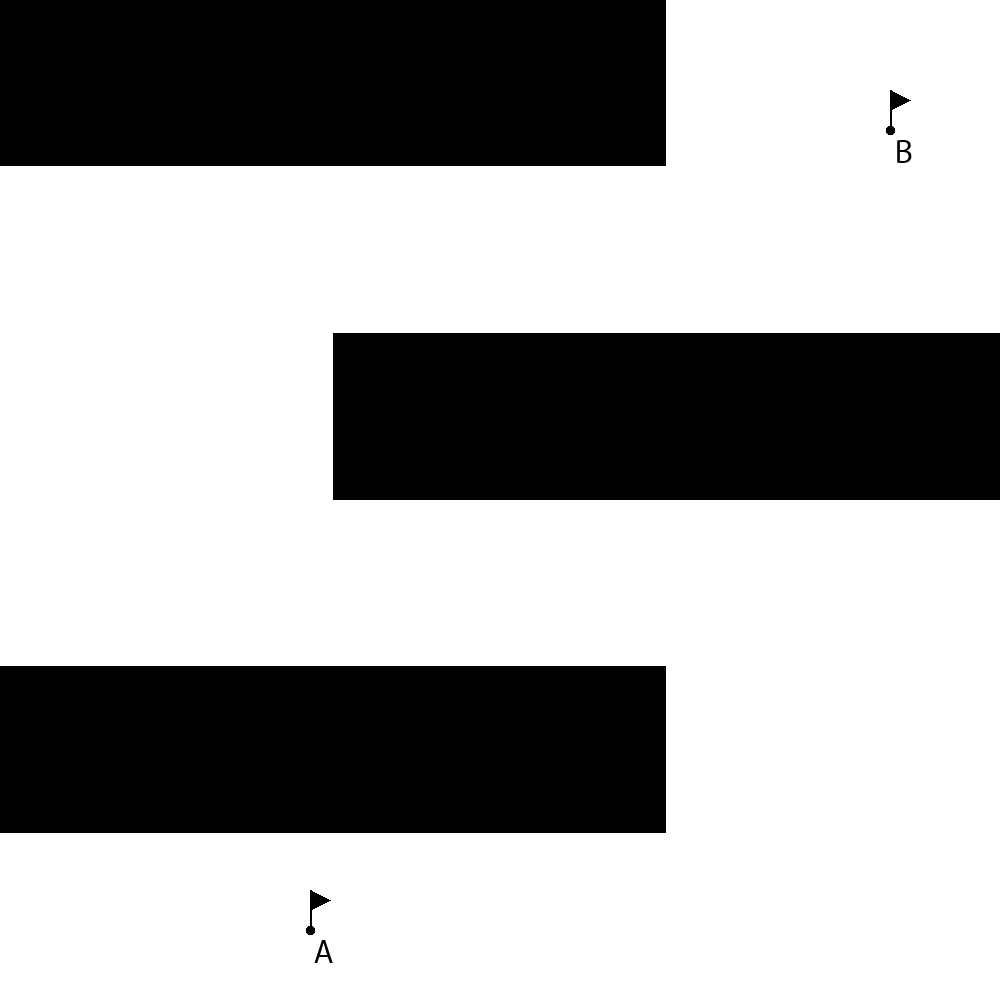}
        \caption{The second simple map.}
        \label{fig:slopes}
    \end{subfigure}
    \hfill 
    \begin{subfigure}[b]{0.3\textwidth}
        \includegraphics[width=4cm]{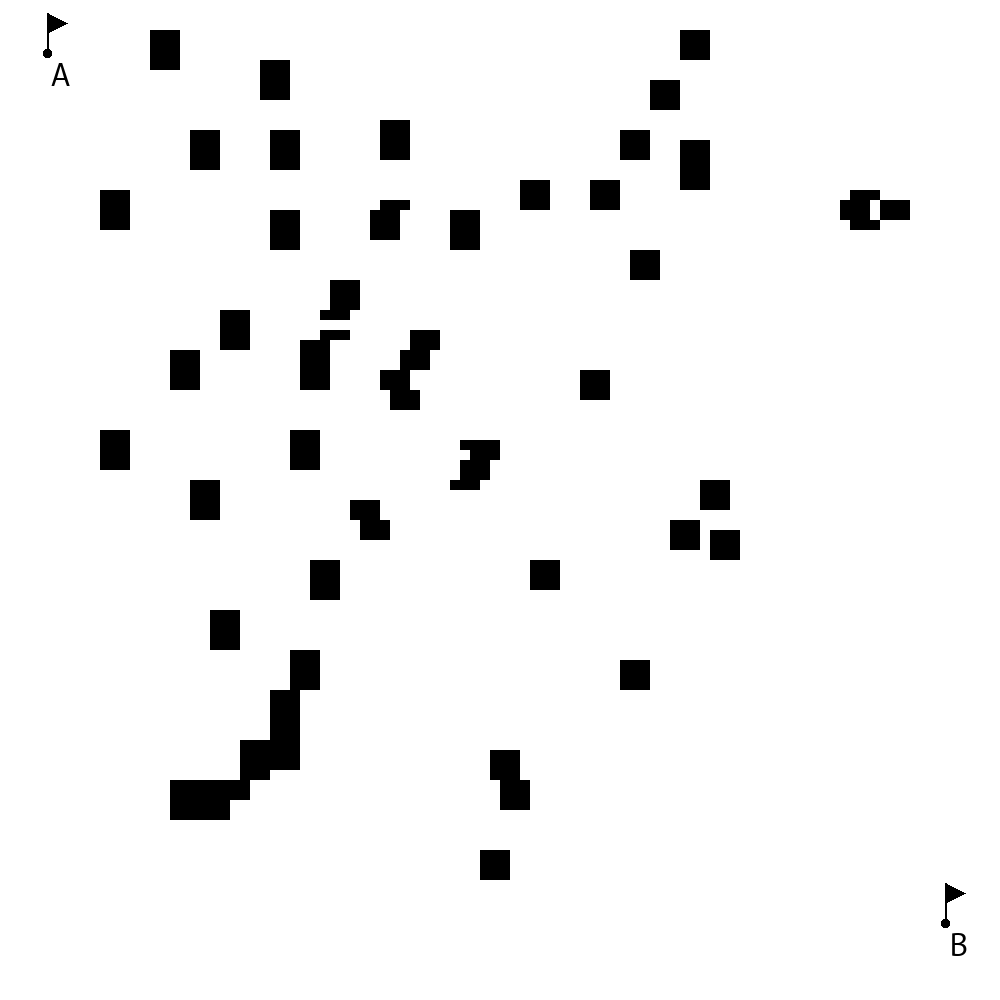}
        \caption{The third (complex) map.}
        \label{fig:islands}
    \end{subfigure}

    \caption{Maps chosen for benchmarking}
\end{figure*}

The map shown in Figure~\ref{fig:move-back-test} is the first small map (it describes an area as large as 0.25km\textsuperscript{2}), that is intended to test how the proposed method is able to handle situations when it is needed to move in a direction that is opposite to the destination. The second small map (it describes an area as large as 0.25km\textsuperscript{2}), shown in Figure~\ref{fig:slopes}, is designed to test whether the proposed method is able to handle the situations when there is a need to change the direction to the opposite one several times in order to pass a kind of labyrinth. The third map, shown in Figure~\ref{fig:islands}, looks like an almost real-life map, it is much larger than the other maps used for testing (it represents an area as large as 400km\textsuperscript{2}) and contains many obstacles that cover a significant area.

In order to demonstrate how the proposed algorithm benefits from parallel genetic algorithms, we have conducted two series of test runs. During the first one, we used only one genetic algorithm to plan the final route, during the second series we used four genetic algorithms running in parallel. For all tests we used personal computer with Intel$^{\tiny{\textregistered}}$ Core i5$^{\tiny{\textregistered}}$ CPU and 8 GiB RAM, this CPU has four logical cores, that is why we used four parallel genetic algorithms. We ran 300 generations of each genetic algorithms, each generation contained 20 individuals, and 10 best individuals of each generation were moved to the next one (elitism), then 10 additional individuals were created using the described algorithm. Mutation probability was set to $0.1$ and crossover probability was set to $0.5$. In case of parallel genetic algorithms, crossover between different islands was performed after every 100 generations; moreover, each genetic algorithm used its own waypoint selection algorithm, as described in section~\ref{sct:planning-route}. Each experiment was run several times in order to tackle biases.

\begin{table*}[h!t]
    \caption{Experimental results.}
    \centering
    \begin{tabular}{| p{2cm} | c | c | c | c | c | c |}
        \hline
        Map & \multicolumn{2}{|c|}{Figure~\ref{fig:move-back-test}} & \multicolumn{2}{|c|}{Figure~\ref{fig:slopes}} & \multicolumn{2}{|c|}{Figure~\ref{fig:islands}} \\
        \hline
        Map size & \multicolumn{2}{|c|}{0.25 km\textsuperscript{2}} & \multicolumn{2}{|c|}{0.25 km\textsuperscript{2}} & \multicolumn{2}{|c|}{400 km\textsuperscript{2}} \\
        \hline
        GA type & non-parallel & parallel & non-parallel & parallel & non-parallel & parallel \\
        \hline
        Min. route length, metres & 858.09 & 835 & 1108.3 & 1072.98 & 25945.8 & 25305.9 \\
        \hline
        Max. route length, metres & 883.13 & 957.32 & 1249.4 & 1171.2 & 27168.4  & 25600.1 \\
        \hline
        Mean route length, metres & 871.63 & 884.94 & 1183.1 & 1121.9 & 26624.68 & 25404.22 \\
        \hline
        Min. time to compute, seconds & 0.409 & 0.386 & 0.397 & 0.478 & 6.033 & 11.084 \\
        \hline
        Max. time to compute, seconds & 0.545 & 0.713 & 0.923 & 0.887 & 8.097 & 12.496 \\
        \hline
        Mean time to compute, seconds & 0.468 & 0.490 & 0.522 & 0.699 & 7.0726 & 11.855 \\
        \hline
    \end{tabular}
    \label{table:result}
\end{table*}

\begin{figure*}[h!t]
    \label{fig:bench-maps-routes}
    \begin{subfigure}[b]{0.3\textwidth}
        \includegraphics[width=4cm]{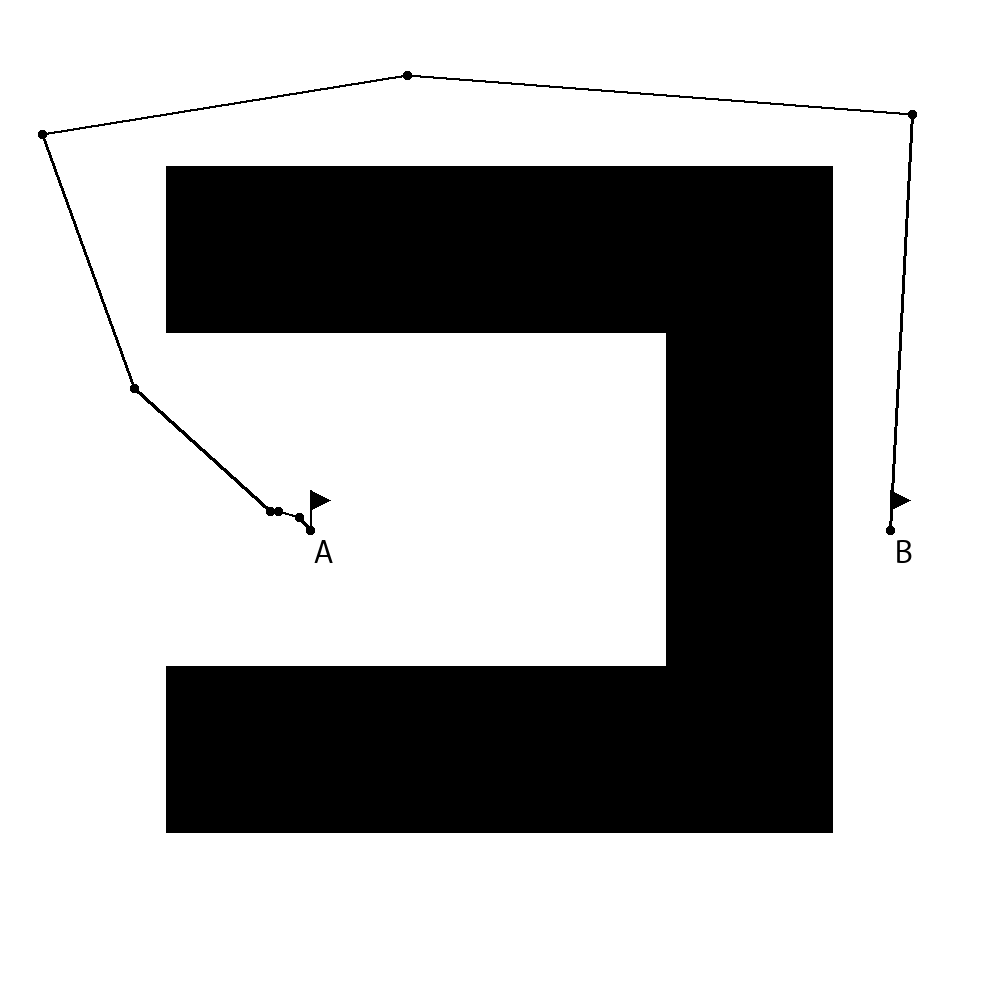}
        \caption{The route for the first simple map.}
        \label{fig:move-back-test-route}
    \end{subfigure}
    \hfill 
    \begin{subfigure}[b]{0.3\textwidth}
        \includegraphics[width=4cm]{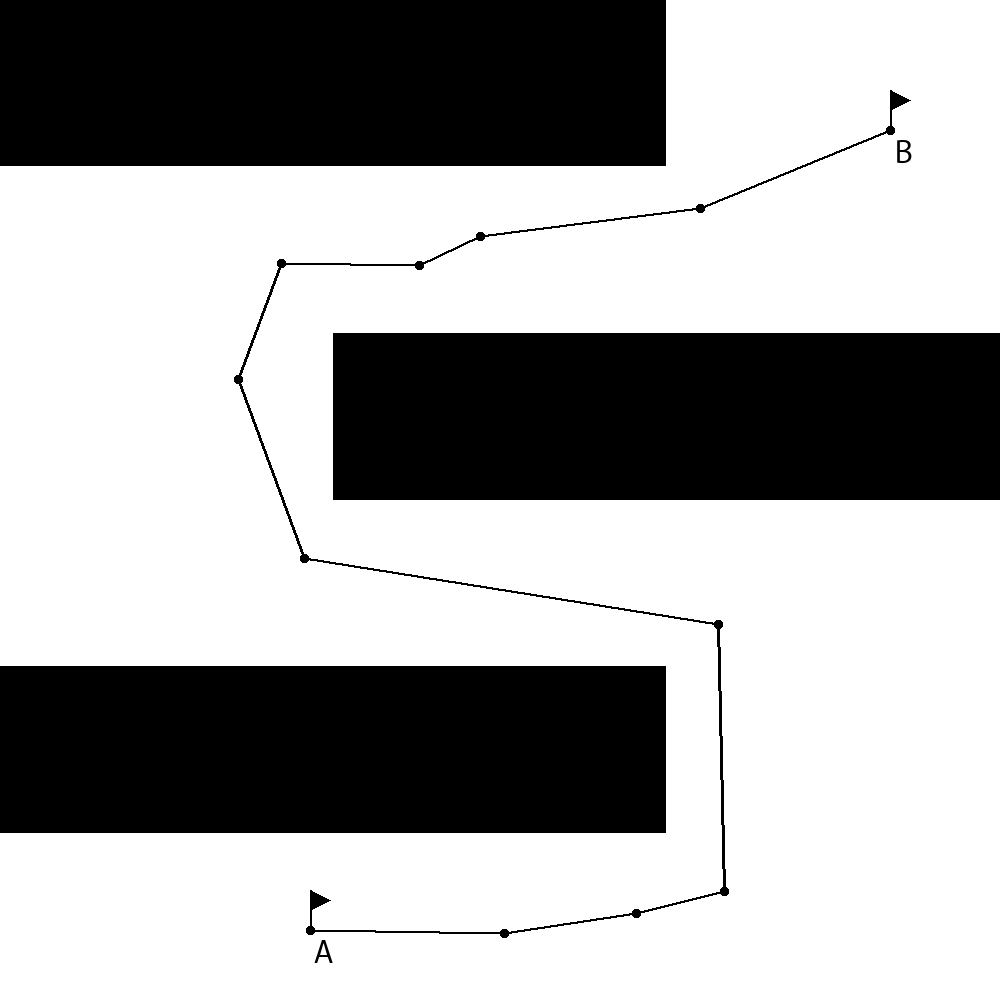}
        \caption{The route for the second simple map.}
        \label{fig:slopes-route}
    \end{subfigure}
    \hfill 
    \begin{subfigure}[b]{0.3\textwidth}
        \includegraphics[width=4cm]{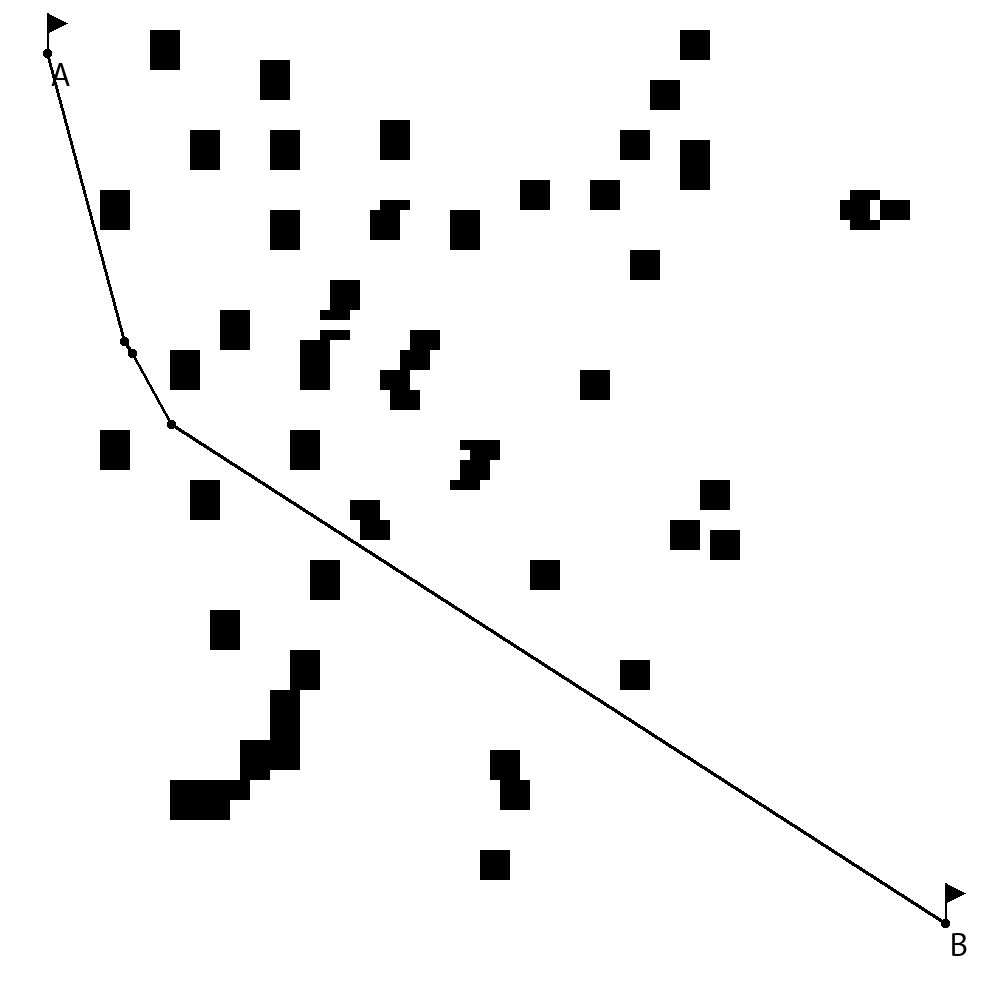}
        \caption{The route for the third (complex) map.}
        \label{fig:islands-route}
    \end{subfigure}

    \caption{Routes created for the selected maps using 4 parallel genetic algorithms.}
\end{figure*}

Table~\ref{table:result} shows experimental results for five runs of each experiment. Figures~\ref{fig:move-back-test-route},~\ref{fig:slopes-route} and~\ref{fig:islands-route} show routes planned by the proposed algorithm for the corresponding maps using four parallel genetic algorithms.

According to the table, in case of a simple map (like~\ref{fig:move-back-test}), the non-parallel genetic algorithm outperforms the parallel one. In case of a more complex, but smaller map (like in Figure~\ref{fig:slopes}), parallel genetic algorithm is able to produce better solutions, but takes slightly more time, which might be due to CPU context switching during multithreaded execution.

In case of a large and complex map (Figure~\ref{fig:islands}), the non-parallel genetic algorithm is able to find a solution faster, but the solution quality is poor: the worst solution produced by a parallel genetic algorithm is comparable to the best solution found by the non-parallel one. Although the parallel genetic algorithm takes more time to find a solution, in absolute numbers the difference is several seconds only. The parallel genetic algorithm is also more stable, the difference between the best and the worst solutions (in terms of length) is much smaller than for the non-parallel one.

Although for more difficult cases or larger maps performance may be lower, we do not think this is an issue. Nowadays industry tends to move as many operations as possible to the on-shore systems, therefore the requested route may be constructed using powerful computational facility located on-shore and then transmitted to the ship~\cite{2015-simonsen-state-of-the-art-weather-routing}.

\begin{figure}
    \centering
    \includegraphics[scale=0.75]{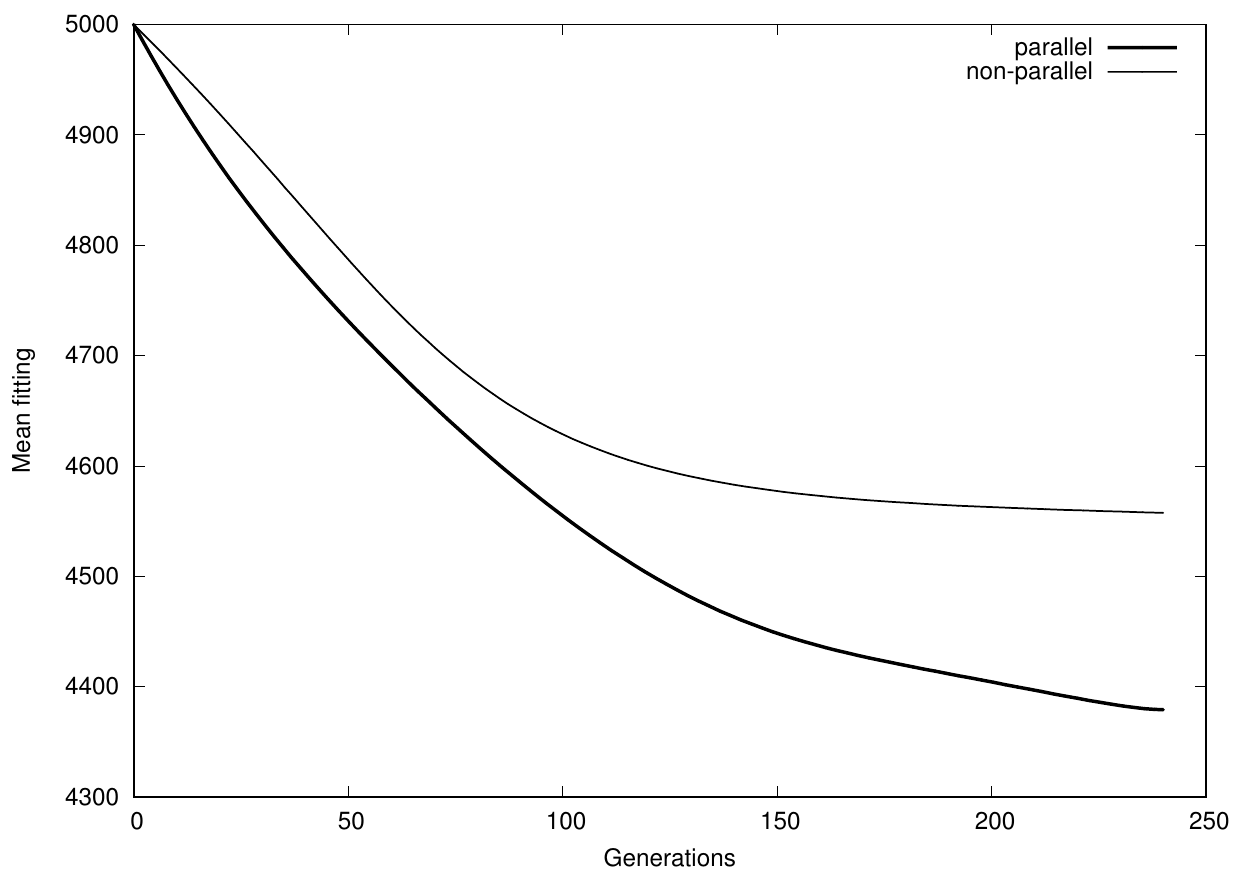}
    \caption{Mean fitting value by generation, lower is better.}
    \label{img:performance-plot}
\end{figure}

The plot in figure~\ref{img:performance-plot} shows how the proposed algorithm benefits from parallel execution in terms of the fitting. This plot shows how mean fitting value changes over time while an algorithm goes through generations. As one can see, the parallel genetic algorithm shows lower mean fitting than the non-parallel one, and the difference becomes more significant with time. Moreover, after the 100th generation mean fitting of the individuals produced by the parallel genetic algorithm decreases dramatically, this is due to the crossover between different islands that is performed every 100 generations. Therefore, the crossover between different islands makes the algorithm even more effective.

\section{Further research}
The described method is capable of planning safe and optimal route for a ship. The method assumes that the landscape is constant, and the depths do not change. This does not reflect the reality: depths change because of tides, and this variation may affect safety of a route. In some cases a ship is able to pass through a particular area only during a tide. In order to take this into account the system that plans routes for ships needs to know when areas are deep enough. Moreover, water currents affect a ship either accelerating it when the direction of a ship is equal to the direction of a current, or reducing the ship's speed otherwise. When planning a route for a ship, it may be essential to know the velocity of currents nearby. When a ship is moving faster than expected, it may perform a manoeuvre at a wrong location, because it was expected to be at a different place, the same is for the situation when a current makes a ship move slower. Finding a way to deal with tides and currents is the subject of the further research.

Waves also affect a ship. In case of relatively open areas, waves may be significantly high, up to several metres. Waves may be so high, that an area becomes too shallow for a ship at trough, being deep enough at crest. Waves reflected from obstacles create local currents that also affect a ship. Additional research is needed in order to determine how to plan a route taking into account waves, depth variations and local currents that waves cause.

As we've already mentioned, ships have complex movement physics involving hydrodynamics. A ship is also able to turn only according a relatively large arc (although the vast majority of modern ships have manoeuvring thrusters, they are usually used only for low-speed manoeuvring, not when a ship is underway). Thus, it is possible to plot such a route that is both safe and optimal, but on the other hand could not be followed by a ship because the route makes a ship perform manoeuvres that are not possible. Tuning the algorithm so that it takes these physical constraints into account is the subject of the further research. We think it could be achieved through modifying the fitting function that examines routes.

\section{Conclusion}
The paper proposes a novel algorithm for ship's route planning. This paper describes a parallel genetic algorithm that can be used to optimize a route between a pair of points and takes advantage of multicore CPUs. This genetic algorithm uses complete routes as individuals. This algorithm uses a novel approach to plan single routes between the start and the destination, which is used to create the initial generation for a genetic algorithm. The proposed algorithm uses heuristic waypoint creation method, that allows to ignore some subareas that are too far from the goal or that are too far from the existing points of a route. Ignoring these parts of the area reduces resource consumption and allows to find the solution faster.

The method does not need a graph that describes the area where the action is taking place, as a result it may be faster and consume less memory than graph-based pathfinding algorithms.

The fitting function of an evolutionary algorithm can apply almost any constraint to a solution, as long as its result could be expressed as a number. We think that this makes the proposed method even more powerful, because it is possible to apply constraints dynamically, turning them on and off according to a particular situation and requirements, making it possible to optimize the route according to several goals.

\end{document}